\PassOptionsToPackage{table,xcdraw,dvipsnames}{xcolor}
\documentclass[journal]{IEEEtran}

\usepackage{amsmath} %
\usepackage{amssymb}  %

\usepackage{graphicx}
\usepackage{subcaption}
\usepackage{booktabs}
\usepackage{multirow}
\usepackage{graphicx}
\usepackage{tcolorbox} 
\usepackage{url}
\definecolor{main}{HTML}{5989cf}    %
\definecolor{sub}{HTML}{f4f9ff}     %

\newtcolorbox{boxB}{
    colback = white,
    boxrule = 1.5pt,
    colframe = main,
    rounded corners,
    arc = 5pt   %
}

\usepackage[capitalize]{cleveref}
\crefname{section}{Sec.}{Secs.}
\crefname{section}{Section}{Sections}
\crefname{table}{Table}{Tables}
\crefname{table}{Tab.}{Tabs.}
\crefname{figure}{Fig.}{Figs.}
\Crefname{figure}{Figure}{Figures}

\title{Peepers \& Pixels:\\ Human Recognition Accuracy on \\ Low Resolution Faces}

\author{
    Xavier Merino$^{1*}$, Gabriella Pangelinan$^{1*}$, Samuel Langborgh$^{1}$, Michael C. King$^{1}$, Kevin W. Bowyer$^{2}$\\
    \begin{minipage}[t]{0.45\textwidth}
        \centering
        $^{1}$Florida Institute of Technology\\
        Melbourne, FL
    \end{minipage}%
    \hfill
    \begin{minipage}[t]{0.45\textwidth}
        \centering
        $^{2}$University of Notre Dame\\
        Notre Dame, IN
    \end{minipage}
}

\begin{document}

\maketitle

\renewcommand{\thefootnote}{\fnsymbol{footnote}} %
\footnotetext[1]{\hspace{-3pt}$^*$Indicates equal contribution.} %
\renewcommand{\thefootnote}{\arabic{footnote}} %

\begin{abstract}
Automated one-to-many (1:N) face recognition is a powerful investigative tool commonly used by law enforcement agencies. In this context, potential matches resulting from automated 1:N recognition are reviewed by human examiners prior to possible use as investigative leads. While automated 1:N recognition can achieve near-perfect accuracy under ideal imaging conditions, operational scenarios may necessitate the use of surveillance imagery, which is often degraded in various quality dimensions. One important quality dimension is image resolution, typically quantified by the number of pixels on the face. The common metric for this is inter-pupillary distance (IPD), which measures the number of pixels between the pupils. Low IPD is known to degrade the accuracy of automated face recognition. However, the threshold IPD for reliability in human face recognition remains undefined. This study aims to explore the boundaries of human recognition accuracy by systematically testing accuracy across a range of IPD values. We find that at low IPDs (10px, 5px), human accuracy is at or below chance levels (50.7\%, 35.9\%), even as confidence in decision-making remains relatively high (77\%, 70.7\%). Our findings indicate that, for low IPD images, human recognition ability could be a limiting factor to overall system accuracy.
\end{abstract}

\section{Introduction}\label{introduction}

Technological systems have become increasingly embedded in both everyday life and the institutions that shape society. While such systems are intended to act as tools to aid in human processes and decision-making, their power and performance has increased rapidly, even surpassing the ability of human experts.
But over-reliance on these systems can have the dangerous side effect of ``automation bias'', in which we begin to place blind trust in 
automated decision-making, and neglect to consider its limitations \cite{MacMillan2025}. The consequences of such over-reliance are especially dangerous in contexts where automated decisions can have drastic real-world impact---such as in the case of law enforcement use of automated face recognition (AFR) systems. To date, there are 
at least seven known cases of wrongful arrest following law enforcement use of AFR \cite{Hill_2024}. In these cases, law enforcement frequently failed to undertake traditional investigative steps that could have eliminated the suspects, e.g., failing to check alibis and collect key evidence \cite{MacMillan2025}.

The first known case of wrongful arrest related to face recognition occurred in 2020 \cite{ACLU_Williams2023}.
Limitations of AFR systems were known at the time. In fact, starting in 2018, various media outlets 
criticized AFR as ``racist'' and ``sexist'' in general \cite{Olson2018, Hoggins2019}, drawing on reports of demographic differentials in accuracy \cite{buolamwini2018gender, raji2020saving, frvt-demographic}.\footnote{Notably, \cite{buolamwini2018gender} and \cite{raji2020saving} actually concerned demographic differentials in gender classification, a form of face \textit{analytics} distinct from face recognition.} Recent media criticism focuses more specifically on the role of AFR in wrongful arrest cases. News stories allege that AFR is ``causing false arrests across [the] nation'' \cite{AtlantaNewsFirst2023} and ``putting innocent people at risk of being incarcerated'' \cite{InnocenceProject2024}. 

Recent research has sought to develop a more nuanced understanding of the limitations of AFR, acknowledging that many systems exhibit performance differentials based not only on demographics but also with respect to image quality issues. However, those who decry the use of face recognition by law enforcement as a single point of failure that directly leads to wrongful arrests fundamentally misunderstand the way that the technology is \textit{meant} to be used---as a tool \textit{supporting} the standard investigative process. 

Investigations that %
use AFR require collaboration between the AFR system and human experts who ``make a series of identity judgments in a predetermined sequence'' \cite{towler2017unfamiliar}. The precondition of using AFR is that an image of a potential suspect exists. This is the probe image. 
Following the guidelines for AFR use outlined in a report from the Major Cities Chiefs Association (MCCA), a human expert should first judge whether the image is suitable for use with AFR \cite{MCCA2021}.
Then, the AFR system compares the probe to a gallery of images of known persons and returns a list of the most likely identity matches. Then, a human expert conducts ``a detailed review of the [returned] imagery...through a manual process of examining the morphological similarities or differences with the unknown subject'' \cite{MCCA2021}. %
The MCCA guidelines suggest that this review step is ``arguably the most important'' \cite{MCCA2021}.

The type of AFR used in this scenario has been shown in experimental settings to have near-perfect accuracy, given high-quality probe and gallery images \cite{grother2019face}. Operationally, however, probes are often taken from surveillance video, which tends to have low quality \cite{zou2011very}. A key quality factor in surveillance images is facial resolution.
Facial resolution is quantified as inter-pupillary distance (IPD), which is the number of pixels between the centers of the pupils. While the minimum IPD required for a given AFR system is algorithm-specific, a recent report from the Organization of Scientific Area Committees for Forensic Science (OSAC) notes that ``FR systems typically work better with a facial image that has between 64 to 128 pixels between the center of the subject’s eyes'' \cite{OSAC2025}. They caution that the lower bound of 64px may be sufficient for ``ideal environmental and operating conditions'', but that non-ideal conditions may require higher IPD \cite{OSAC2025}. Low IPD images can significantly degrade performance, even for state-of-the-art recognition systems \cite{pangelinan2024analyzing, bhatta2023demographic}. 

In the law enforcement use case, the limitations of the AFR component are clear, with well-defined image quality thresholds. But what about the limitations of the \textit{human(s)} involved in the process? Relatively little work has systematically studied the impact of low resolution on human recognition ability, and to our knowledge, none have clearly defined a threshold for its reliability.

We design and implement an experiment to answer this motivating question. Subjects are asked to make identity judgments for pairs of face images consisting of one high-IPD ``reference image'' and one low-IPD ``probe image''. 
This is meant to mimic the situation of comparing a low-quality probe image with a high-quality gallery image returned by the AFR system.
We vary the IPD of the probe image from a maximum of 30px to a minimum of 5px. By examining accuracy at systematically reduced IPD values, we can develop a more precise understanding of the threshold of human recognition accuracy.

The paper is organized as follows. Section \ref{background} gives an overview of the general law enforcement face recognition process, contextualizing the role of image quality and types of recognition involved. Section \ref{relatedwork} summarizes findings from related work, providing an understanding of baseline human accuracy in ideal and non-ideal conditions. Sections \ref{data-prep} and \ref{experiment-design} describe the design and implementation of the test. Section \ref{results} presents and analyzes the experimental results. Section \ref{conclusion} summarizes our findings and discusses the implications.

\section{Background}\label{background}

To provide context for our work, we first present an overview of how face recognition appears to typically be employed %
in criminal investigations. The process can vary by jurisdiction, and this overview may not necessarily represent the exact steps and tools used by any given agency. However, we can refer to the case of Williams v. City of Detroit, which was both the ``first publicly reported instance of a false face-recognition `match' leading to a person's wrongful arrest'', and the first such case for which the resulting lawsuit has been settled in court \cite{ACLU_Williams2023}. As the associated legal documents are now publicly available, we can look to them to provide a general idea of the face recognition process in investigation and concrete examples of each step. 

\subsection{Example FR Pipeline}
The law enforcement FR pipeline begins when a crime occurs and its suspect is captured in an image or video. 
In the Williams case, an unknown individual was recorded on video surveillance committing theft \cite{King2023}.

\begin{boxB}
An image of the suspect is extracted from video footage. It is referred to as the \textbf{\textcolor{main}{probe image}}.
\end{boxB}
In the Williams case, the surveillance camera yielded three still images taken from video frames. Each image had ``low pixel density'' \cite{King2023}. The MCCA Report states that an evaluation of whether ``the imagery is appropriate for use'' in a given agency's AFR system ``should apply at least in part to any agency's program'' \cite{MCCA2021}.  In the Williams case, the image quality was assessed as ``poor'' \cite{King2023}.
However, the image was then submitted to the AFR system.

\begin{boxB}
The probe is input to an \textbf{\textcolor{main}{FR system}}.
\end{boxB}
In the Williams case, the DataWorks Plus system was used, which integrated two commercial FR engines: Rank One Computing (ROC) \cite{roc_ai} and NEC Corporation \cite{nec_biometrics} \cite{King2023}. Based on the testimony, the Michigan State Police image analyst appears to have been more reliant on the candidate list produced by the ROC algorithm, as its version release date was circa 2018, while the NEC's was 2011.

\begin{boxB}
The FR system compares the probe to every face image in the \textbf{\textcolor{main}{gallery}}, a database of N images with labeled identities.
\end{boxB}
In the Williams case, the gallery contained ``approximately 49 million images, including current and expired Michigan driver’s license and state ID photos [and] arrest photos from law enforcement agencies across the state'' \cite{King2023}.

\begin{boxB}
When the probe is compared to each individual gallery image, the FR engine generates a score indicating their similarity. The most similar gallery images are output as the \textbf{\textcolor{main}{candidate list}}.
\end{boxB}
In the Williams case, the lists returned by the two AFR algorithms each had 243 potential matches (486 total). An expired driver's license image of Williams was in the ninth position in the ROC list.
Notably, his then-current license image, which was also in the gallery, did not appear in the list. Neither image appeared in the NEC list~\cite{King2023}. 

\begin{boxB}
An image analyst reviews the candidate list and selects the image they believe is the most likely identity match to the probe. The selected image is the \textbf{\textcolor{main}{investigative lead}}.
\end{boxB}
In the Williams case, the image analyst reported being able to ``identify several features that were consistent between the probe image and the image of Mr. Williams'', despite also noting that the probe image was of poor quality \cite{King2023}.

\subsection{1:N and 1:1 Matching}

The law enforcement face recognition pipeline features the two types of face recognition. 

The first type is 1:N matching (``identification''), and it occurs when the FR engine compares the probe image to the N gallery images. The image examiner also performs 1:N matching when comparing the probe image to the images of the candidate list, albeit for significantly smaller N. As we are primarily interested in the human element, we will discuss 1:N matching with respect to the second instance.

Given a probe image with unknown identity and a gallery of images with labeled identities, the fundamental task of 1:N matching is to find the most likely identity match. If the investigative lead selected by the examiner is the correct identity of the probe, they have made a \textit{true positive identification}. If it is not, they have made a \textit{false positive identification}. If the correct identity exists in the gallery but is \textit{not} selected, they have made a \textit{false negative identification}. The latter two errors are quantified as False Positive and False Negative Identification Rates (FPIR and FNIR).

The second type of recognition is 1:1 matching (``verification''), and it is performed by the FR system for each (probe, gallery image) pairing. A human may also perform 1:1 matching, e.g., if a secondary examiner compares the probe \textit{directly} to the investigative lead, without examining the  candidate list.

Given a probe image with unknown identity and a second image, the fundamental task of 1:1 matching is to determine whether the two images show the same person or different persons. Two images of the same person comprise a mated pair, and failure to classify the pair as such is referred to as a \textit{false non-match}. Two images of different persons comprise a non-mated pair, and failure to classify the pair as such is referred to as a \textit{false match}. These errors are quantified as False Non-Match and False Match Rates (FNMR and FMR).

The image quality factors that can cause AFR systems to yield increased error rates are generally well-understood. Images featuring overly light, dark, or uneven illumination, for example, can increase FMR and FNMR \cite{wu2023face,pangelinan2025lights}. If the probe has significant blur or very low resolution, FPIR can increase \cite{pangelinan2024analyzing}. These are the very same factors that tend to be present in surveillance imagery. Thus, our work is broadly rooted in understanding how these quality issues impact the humans performing matching in the law enforcement case--- beginning here with the quality factor of resolution.

\section{Related Work}\label{relatedwork}

Many previous works that evaluate human accuracy on identity-comparison tests come from the fields of psychology and cognitive science. Those in the computational realm often have the goal of comparing or combining human and algorithmic performance. As such, we must carefully discern which studies are preferable for informing our test design and implementation and providing comparable accuracy results. We review works fulfilling the following criteria:
\vspace{0.5em}
\begin{itemize}
    \item Evaluation of 1:1 rather than 1:N performance \cite{wilkinson2009facial, white2015error}
        \item Inclusion of untrained participants
    \begin{itemize}
        \item We want to understand baseline human ability on low-quality images before evaluating experts.
    \end{itemize}
    \item Interpretation of accuracy based on individual rather than combined decisions \cite{kumar2011describable, Best2014unconstrained}
    \begin{itemize}
        \item Fused judgments are significantly more accurate than individual \cite{towler2017unfamiliar, white2013crowd, phillips2015perceptual, white2015perceptual}, and we find no evidence to indicate fusing is used operationally.
    \end{itemize}

\end{itemize}

The works discussed in Secs. \ref{relatedwork-baseline} and \ref{relatedwork-lowres} present tests that consist of paired images. In most cases, reviewers are asked to rate their confidence on a Likert scale, where the lowest value indicates strong confidence that the images depict different identities, the highest value indicates strong confidence that they depict the same identity, and the midpoint indicates uncertainty or inability to decide. 

\subsection{Human Accuracy Baseline}\label{relatedwork-baseline}

For a baseline of human accuracy in \textit{ideal} matching conditions, we refer to the Glasgow Face Matching Test (GFMT) designed by Burton et al. \cite{burton2010glasgow}. The test consisted of 168 grayscale image pairs (84 mated, 84 non-mated). The identities appearing in the pairs were not distinct, and non-mated pairs were selected for difficulty based on human perception. The test was presented in-person to 300 untrained reviewers, and they achieved a mean accuracy of 89.9\%, with slightly higher performance on mated (92\%) than non-mated pairs (88\%). Interestingly, the key motivation of this work was presenting pairs of images taken on different cameras, as ``the issue of camera change is an important one in forensic settings'' \cite{burton2010glasgow}. Yet all the images represented ideal recognition conditions: high-quality, front-facing, and well-illuminated. Additionally, the images in mated pairs were captured in a single session, essentially avoiding any variations in appearance. For these reasons, the 89.9\% accuracy achieved by humans in the GFMT may be understood as a best-possible accuracy achieved under idealized conditions. 

The accuracies reported in \cite{fysh2018kent, phillips2018face, jeckeln2024designing} provide a more realistic baseline, testing human ability on images which are not necessarily ideal but are generally of high quality.
Fysh and Bindemann \cite{fysh2018kent} designed the Kent Face Matching Test (KFMT) to address the real-world limitations of the GFMT by introducing variations in pose, illumination, and expression (PIE). Pairs featured one passport-style image and one ``ambient'' student ID image (with less-controlled PIE, though the images were still captured indoors and subjects were cooperative), with an average of 8.8 months between acquisitions. The short version of the KFMT consisted of 40 image pairs (20 mated, 20 non-mated) of the Caucasian demographic. Non-mated pairs were manually selected to be difficult based on visual similarities perceived by the researchers. The test was presented in-person to 60 untrained reviewers, and they achieved a mean accuracy of 66\% for both mated and non-mated pairs. 

 Phillips et al. \cite{phillips2018face} designed a test consisting of 20 image pairs (12 mated, 8 non-mated) selected for difficulty (i.e. mated / non-mated pairs assigned low / high similarity scores by an FR algorithm). Pairs featured one indoor image and one outdoor image ``with limited control of illumination, expression, and appearance'' \cite{phillips2018face}, and image subjects were primarily of the Caucasian male demographic. The test was presented in-person to 31 untrained reviewers, and they achieved a median accuracy of 68\%.

 Jeckeln et al. \cite{jeckeln2024designing} designed a cross-race test consisting of 20 image pairs (10 mated, 10 non-mated), equally distributed across race (Black, White) and selected for difficulty as in \cite{phillips2018face}. The test was presented in-person to 105 untrained reviewers (approximately half Black, half White), and median accuracy for both groups ranged from 56-64\%.

\subsection{Human Accuracy on Low Resolution Images}\label{relatedwork-lowres}
 
Norell et al. \cite{norell2015effect} designed a test consisting of 30 image pairs, where each pair contained one high-quality (``reference'') image taken in a studio setting and one varying-quality (``questioned'') image taken from surveillance footage. The questioned images had three possible quality levels, corresponding to resolutions of 50x68, 86x114, or 286x386 for the full images. The test was presented to 67 total participants, consisting of 17 trained reviewers (forensic experts) and 50 untrained reviewers (students). Responses were reported on a -4 to +4 scale. 

In analyzing their results, Norell et al. first examined whether better image quality yielded a more certain conclusion (e.g., -4 vs. -2). (Responses of 0 indicating uncertainty were excluded.) For trained reviewers, image quality was found to be highly correlated with certainty, particularly for mated image pairs. Quality also significantly affected their accuracy, decreasing consistently from 94-99\% for highest-quality pairs to 45-54\% for lowest-quality pairs. Untrained reviewers had the lowest accuracy for the lowest-quality pairs, although this was \textit{not} consistent across quality levels---for mated pairs, they actually had higher accuracy on the second-highest quality images than the highest-quality. 

Keval and Sasse \cite{keval2008can} designed a test of 64 image pairs containing one high-quality image and one low-quality CCTV still, corresponding to four video quality bit rates (32, 52, 72, and 92 Kbps). The test was presented to 80 untrained participants, who provided identity decisions of Yes or No. They found that participant accuracy decreased $\sim$6-18\% as video quality decreased. They also qualitatively evaluated participant feedback, drawing the conclusion that for lowest-quality images, participants found that internal facial features were less clear (particularly for image subjects with darker skin tones) and thus relied primarily on external features. 

Bindemann et al. \cite{bindemann2013effect} designed a test of 160 image pairs containing one high-quality image and one pixelated image, corresponding to 20, 14, or 8 pixels across the full face. All images were derived from the GFMT dataset \cite{burton2010glasgow}, and the pixelated versions were generated by combining groups of multiple pixels into single larger pixels. The test was presented to 20 untrained reviewers, who provided identity decisions of Yes / No. Accuracy ranged from 48-66\% for mated pairs and was $\sim$60\% across the three pixelation levels.  

Our work differs from the previous three in several key aspects. (1) Resolution in \cite{norell2015effect, keval2008can} is reported in terms of the full images, rather than IPD, and therefore the resolution of the face region itself is not specified.\footnote{In \cite{bindemann2013effect}, vertical / horizontal resolution of the full face is given, but this is not the commonly used metric.} By specifying IPD directly, we can systematically vary its value in order to determine the threshold of human accuracy. (2) We employ a Likert scale for responses, which reflects how forensic examiners in operational scenarios make their decisions \cite{phillips2018face, norell2015effect, Hahn2022}, rather than a Yes / No response \cite{keval2008can, bindemann2013effect}. (3) We specify accuracy on mated / non-mated pairs separately, rather than overall accuracy alone \cite{norell2015effect}.
This is important for the operational use case, in which false matches / non-matches made by image examiners may lead to wrongful arrest / failure to investigate the right person. (4) We use downsampling to simulate low-resolution imaging conditions, as is common practice \cite{peng2019low}, rather than pixelation \cite{bindemann2013effect}. (5) Finally, we use a larger number of image pairs (320 versus 30-160) and participants (100 versus 20-80).

\section{Data Preparation}\label{data-prep}

\subsection{Dataset}\label{dataset}
For image datasets used in face recognition, variations in pose, illumination, and expression can substantially impact accuracy \cite{Jain-AFR-Study}. Thus, datasets with controlled imaging conditions (i.e. featuring a frontal pose, neutral expression, and standard illumination) are preferred \cite{Jain-AFR-Study}. 

The MORPH dataset \cite{ricanek2006morph} is a popular choice among face recognition researchers, notable for its size and standardization. It contains mugshot images taken in controlled conditions and annotated with demographic information. 
We use the version of MORPH curated in \cite{gender-differences}. It includes 35,276 images of 8,835 Caucasian males (with an average of 4 images per identity) and 56,245 images of 8,839 African American males (average of 6 per identity). 

\begin{figure*}[!ht]
    \centering
    \begin{subfigure}{0.16\textwidth}
        \centering
        \includegraphics[width=\linewidth]{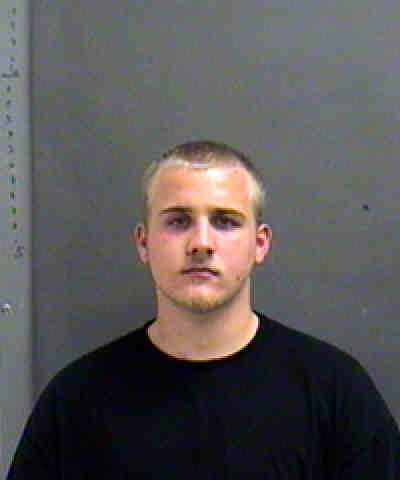}
        \caption{Low IPD}
        \label{ex-lowipd}
    \end{subfigure}
    \begin{subfigure}{0.16\textwidth}
        \centering
        \includegraphics[width=\linewidth]{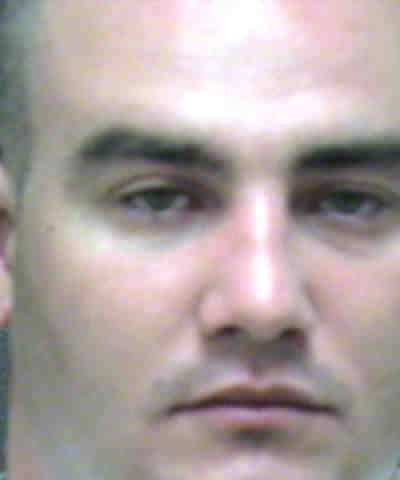}
        \caption{High IPD}
        \label{ex-highipd}
    \end{subfigure}
    \begin{subfigure}{0.16\textwidth}
        \centering
        \includegraphics[width=\linewidth]{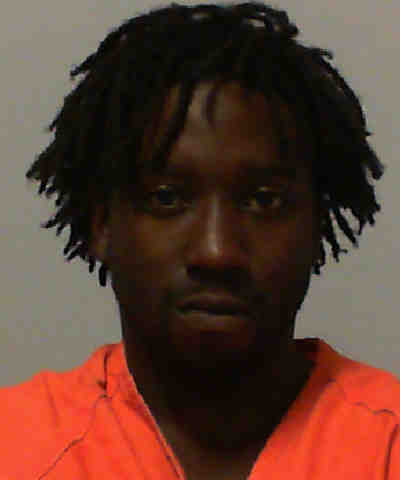}
        \caption{Underexposed}
        \label{ex-under}
    \end{subfigure}
    \begin{subfigure}{0.16\textwidth}
        \centering
        \includegraphics[width=\linewidth]{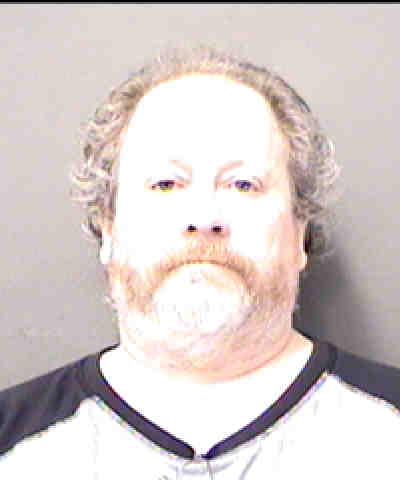}
        \caption{Overexposed}
        \label{ex-over}
    \end{subfigure}
    \begin{subfigure}{0.16\textwidth}
        \centering
        \includegraphics[width=\linewidth]{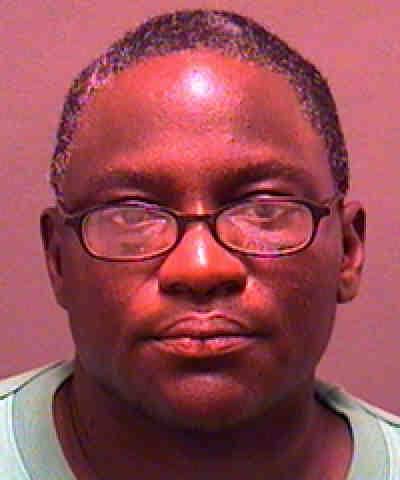}
        \caption{Color Cast}
        \label{ex-color}
    \end{subfigure}
    \begin{subfigure}{0.16\textwidth}
        \centering
        \includegraphics[width=\linewidth]{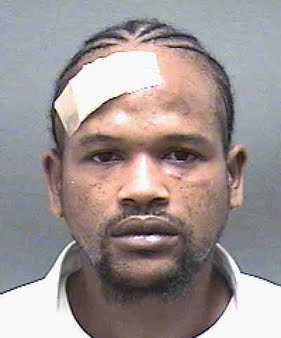}
        \caption{Occlusion}
        \label{ex-occluded}
    \end{subfigure}

    \caption{Ex. discarded images.}
    \label{fig:ex-removed-images}
\end{figure*}

\subsection{Standardizing IPD}\label{standardization}

First, we calculate IPD for all MORPH images. We take the interquartile range (IQR) of the IPD distribution as the ``acceptable'' range: $[92, 110]$ with a mean of 101 and standard deviation of 5.4. Images with IPD values outside of this range tend to represent non-standard subject positioning, e.g., too far from or too close to the camera, as shown in Figs. \ref{ex-lowipd} and \ref{ex-highipd}.

\subsection{Standardizing Brightness}
Next, using the method described in \cite{wu2023face}, we calculate brightness for all images in the $L^{*}a^{*}b^{*}$ colorspace. Using the $L^{*}$ (perceptual lightness) component as a measure of brightness, we determine the acceptable brightness range to be $[51, 77]$, corresponding to the 2nd through 4th quintiles of the distribution. We perform this step to minimize any impact on accuracy related to illumination effects. Images with brightness values outside of the acceptable range are shown in Figs. \ref{ex-under} and \ref{ex-over}.

Once we have determined the acceptable IPD and brightness ranges for the entire dataset, we turn to our two cohorts of interest: Caucasian and African American Males. Within each of these cohorts, we discard images outside of either range. The intersection of the remaining images---those which have both acceptable brightness and acceptable IPD---is the pool we further work with. 

\subsection{Removing Unsuitable Images}

As a final quality check, we perform a manual review to discard any images deemed unsuitable for the experiment. ``Unsuitable'' may be due to image acquisition conditions (``image-based'' factors) or subject behavior / appearance (``subject-based'' factors). While the MORPH dataset is generally standardized, some image-based factors remain---e.g., abnormal coloration, as in Fig. \ref{ex-color}. The main subject-based factors that led to image removal were non-standard pose and facial obstruction, as in Fig. \ref{ex-occluded}.

\subsection{Selecting Difficult Pairs}\label{diff-pairs}

Now that we have a ``standardized subset'' of African American and Caucasian male images, we can generate similarity scores within each cohort. For matching, we use a combined margin model based on ArcFace loss \cite{arcface} trained on Glint360K(R100)---``the largest and cleanest face recognition dataset'' \cite{glint360k}---with weights from \cite{arcface-github}. For each image pair, the matcher (``ArcFace'') provides similarity scores from -1 to 1, with higher values indicating greater similarity. 
We use the scoring information to select difficult pairs: non-mated (different-person) pairs with high similarity scores and mated (same-person) pairs with low similarity scores. This selection simulates the difficulty level of images that an AFR system returns in the candidate list. 

We next require that each identity is shown in at most one pair. An identity that appears in an non-mated pair appears exactly once in the entire study. An identity that appears in a mated pair appears exactly twice (in those two images) in the entire study. The final set of pairs is equally balanced across Caucasian / African American subjects and mated / non-mated pairs.

\section{Experiment Design}\label{experiment-design}

\subsection{Impression Generation}
We prepare the final set of image pairs for the experiment as follows. Each pair will contain one high-IPD image (``reference image'') and one low-IPD image (``probe image''). We refer to this pairing of a reference / probe image, as presented to study participants, as an ``impression''. The reference image is shown with its original 400x480 resolution. The probe image is generated by first downsampling the full-resolution image with Lanczos resampling (using \cite{clark2015pillow}) then upscaling using bicubic interpolation, so that the two images can be shown as the same size. 

In selecting what values to use for the low-IPD images, we refer to the work of Peng et. al \cite{peng2019low}, which presents a ``resolution scale'' based on biometric standards. They determine that the barrier between high and low resolution occurs at an IPD of 50px, in accordance with the ISO/IEC 19794-5:2005 \cite{ISO19794-5:2011} and ANSI/INCITS 385-2004 \cite{ANSI} standards. The low resolution range is further subdivided into three categories: (1) ``upper low resolution (ULR)'' for IPDs of 50-25px, as 25px is the minimum specified by the EN 50132-7 standard (which relates to CCTV footage), (2) ``moderately low resolution (MLR)'' for 25-13px, and (3) ``very low resolution (VLR)'' for 13-0px. Thus, we select one IPD value in the ULR range to act as the baseline (30px), two in the MLR range (20px, 15px), and two in the VLR range (10px, 5px). 

\begin{figure*}[h]
    \centering
    \begin{subfigure}{0.16\textwidth}
        \centering
        \includegraphics[width=\linewidth]{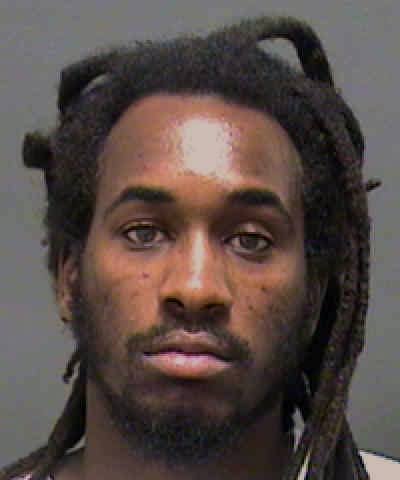}
        \caption{Original}
    \end{subfigure}
    \begin{subfigure}{0.16\textwidth}
        \centering
        \includegraphics[width=\linewidth]{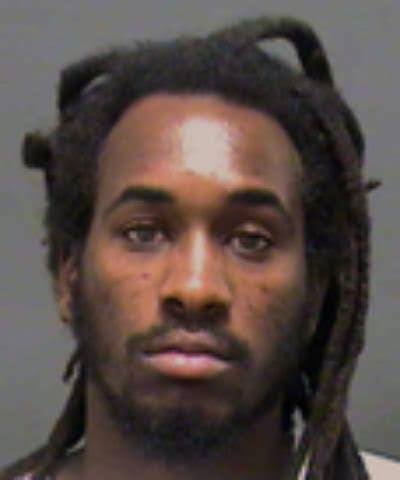}
        \caption{30px}
    \end{subfigure}
    \begin{subfigure}{0.16\textwidth}
        \centering
        \includegraphics[width=\linewidth]{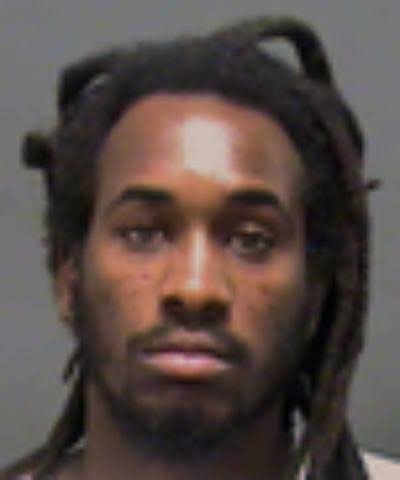}
        \caption{20px}
    \end{subfigure}
    \begin{subfigure}{0.16\textwidth}
        \centering
        \includegraphics[width=\linewidth]{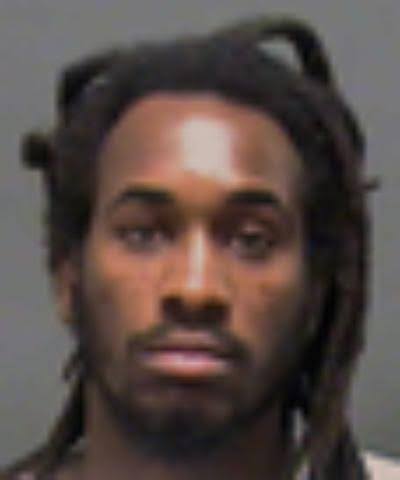}
        \caption{15px}
    \end{subfigure}
    \begin{subfigure}{0.16\textwidth}
        \centering
        \includegraphics[width=\linewidth]{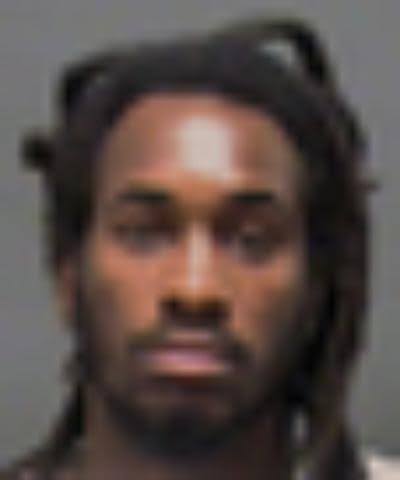}
        \caption{10px}
    \end{subfigure}
    \begin{subfigure}{0.16\textwidth}
        \centering
        \includegraphics[width=\linewidth]{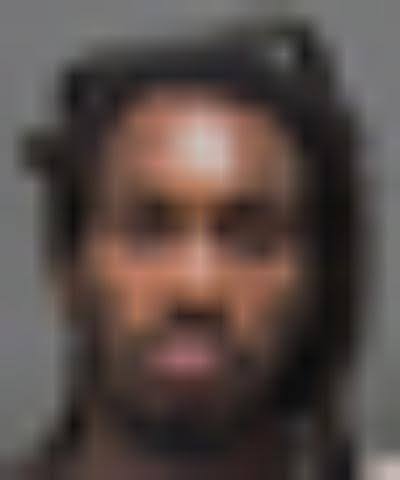}
        \caption{5px}
    \end{subfigure}

    \caption{Visualizing an original image with $\sim$100px IPD to low-IPD versions.}
    \label{fig:ipd-comparison}
\end{figure*}

The five IPD levels are balanced evenly across the probe images, and the corresponding impressions are the same for all participants. For an illustration of how an image looks at the different IPD levels, see Fig. \ref{fig:ipd-comparison}. 

\subsection{Participant Selection}
All 100 participants in the experiment were untrained on the task, and were either university students or employees. We balanced demographic diversity to the extent possible, with the following distributions as self-reported by participants:
\begin{itemize}
    \item Gender: Female (50), Male (47), Other (3)
    \item Race: White (35), Asian (32), Hispanic (17), Black (12), Other (4)
    \item Age: $<$ 20 (25), 20-30 (61), 30-40 (8), 40-50 (1), 50-60 (3), 60+ (2)  
\end{itemize}

\subsection{Test Procedure}

The test consisted of 320 total impressions. The impressions were balanced evenly across IPD of the probe image (64 impressions per IPD), type of pair (32 mated, 32 non-mated), and race of subject (16 African American, 16 Caucasian). The impressions seen by participants are the  same (i.e., always the same pairing of full-resolution reference image and low-IPD probe image) and presented in the same order.

In addition, 16 ``attention checks'' were presented at regular intervals (the same for all participants) to verify engagement. Each attention check consisted of the exact same high-quality image being shown in the positions of the reference image and probe image. 

 Each impression was shown for 15s total, with a 1s delay before the participant was able to respond to the prompt: \textit{Determine whether the images show the same person or not.} The response choices were: \textit{No, Probably Not, Unsure, Probably Yes, Yes}. If the participant did not respond within the time frame, the response was recorded as \textit{Unsure}. The next impression was shown immediately after a timeout or response selection.

\section{Results \& Analysis}\label{results}

\begin{figure*}[!ht]
    \centering
    \includegraphics[width=.8\textwidth]{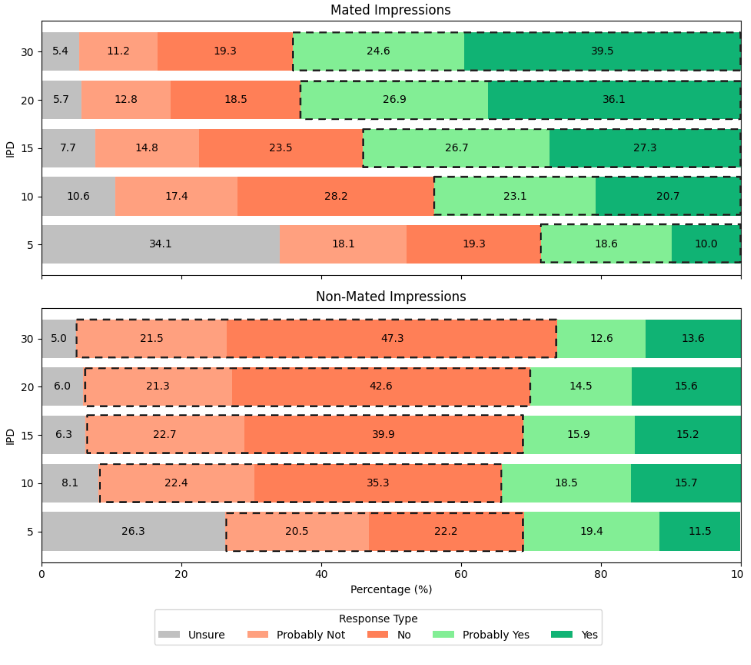}
    \caption{Distribution of responses with respect to IPD level, with correct responses based on pair type outlined.}
    \label{fig:response-distribution}
\end{figure*}

Fig. \ref{fig:response-distribution} shows the distribution of participant responses, separated by IPD value of the probe for each impression type (mated / non-mated). Correct responses for mated / non-mated impressions are outlined in black. Responses of \textit{Probably Not / Probably Yes} are used to gauge certainty, and are interpreted as \textit{No / Yes} in terms of correctness. 

Our main objective is to investigate the impact of IPD on response accuracy. We also explore its influence on decision-making certainty and response time. Our analysis first considers all impressions collectively before distinguishing between mated and non-mated impressions.

\subsection{Results for All Impressions}
Tab. \ref{tab:overall-accuracy} summarizes the overall results for each IPD value.

\begin{table}[!ht]
\centering
\caption{Overall results by IPD value of probe.}
\label{tab:overall-accuracy}
\resizebox{\columnwidth}{!}{%
\begin{tabular}{cc|cc|cc}
\cline{3-4}
\multicolumn{1}{l}{} & \multicolumn{1}{l|}{} & \multicolumn{2}{c|}{\textbf{\% Change in Acc.}} & \multicolumn{1}{l}{} & \multicolumn{1}{l}{} \\ \hline
\multicolumn{1}{|c|}{\textbf{IPD}} & \textbf{\begin{tabular}[c]{@{}c@{}}Acc.\\ (\%)\end{tabular}} & \multicolumn{1}{c|}{\textbf{\begin{tabular}[c]{@{}c@{}}vs. \\ Prev.\end{tabular}}} & \textbf{\begin{tabular}[c]{@{}c@{}}vs. \\ Base.\end{tabular}} & \multicolumn{1}{c|}{\textbf{\begin{tabular}[c]{@{}c@{}}Cert.\\ (\%)\end{tabular}}} & \multicolumn{1}{c|}{\textbf{\begin{tabular}[c]{@{}c@{}}Time\\ (s)\end{tabular}}} \\ \hline
\multicolumn{1}{|c|}{30} & 66.2 & \multicolumn{1}{l|}{} & \multicolumn{1}{l|}{} & \multicolumn{1}{c|}{79.3} & \multicolumn{1}{c|}{5.79} \\ \hline
\multicolumn{1}{|c|}{20} & 63.5 & \multicolumn{1}{c|}{-4.0} & -4.0 & \multicolumn{1}{c|}{78.5} & \multicolumn{1}{c|}{5.75} \\ \hline
\multicolumn{1}{|c|}{15} & 58.3 & \multicolumn{1}{c|}{-8.2} & -11.9 & \multicolumn{1}{c|}{78.0} & \multicolumn{1}{c|}{5.72} \\ \hline
\multicolumn{1}{|c|}{10} & 50.7 & \multicolumn{1}{c|}{-13.0} & -23.4 & \multicolumn{1}{c|}{77.0} & \multicolumn{1}{c|}{4.58} \\ \hline
\multicolumn{1}{|c|}{5} & 35.9 & \multicolumn{1}{c|}{-29.3} & -45.9 & \multicolumn{1}{c|}{70.7} & \multicolumn{1}{c|}{4.52} \\ \hline
\end{tabular}%
}
\end{table}

\subsubsection{Impact of IPD on Overall Accuracy}
Tab. \ref{tab:overall-accuracy} provides the overall accuracy percentage for each IPD level (\textit{Acc.}), as well as the percentage change in accuracy incrementally for each level versus the previous level (\textit{\% Change in Acc. vs. Prev.}) and versus the baseline IPD of 30px (\textit{\% Change in Acc. vs. Base.}). 

Accuracy decreased consistently as IPD decreased. Participants were 46\% less accurate for image pairs with a 5px probe image than for pairs with a 30px probe image. 

For IPDs of 30-15px, accuracy was above chance, ranging from 66.2-58.3\%. However, at an IPD of 10px, accuracy was essentially at chance level (50.7\%), and at an IPD of 5px, accuracy was \textit{below} chance (35.9\%). 

\subsubsection{Impact of IPD on Overall Certainty}
Tab. \ref{tab:overall-accuracy} provides certainty percentage (\textit{Cert.}) for responses at each IPD level. Certainty is calculated using Eq. \ref{c_score}:

\begin{equation}\label{c_score}
    Certainty = \frac{\Sigma_{r}n_{r} \cdot w_{r}}{\Sigma_rn_{r}}
\end{equation}

where $r$ represents a response selection $r \in$ \{No, Prob. Not, Unsure, Prob. Yes, Yes\}, with corresponding weights $w \in$ \{1, 0.5, 0, 0.5, 1\}. 

Accuracy and certainty exhibited a strong positive correlation, with a Pearson correlation coefficient of approximately 0.96. Notably, although accuracy was generally low and fluctuated considerably across IPD levels (36\%–66\%), certainty remained relatively high and consistent (71\%–79\%). This suggests that participants were generally confident in their responses, even when their accuracy was poor.

\subsubsection{Impact of IPD on Overall Time}

Tab. \ref{tab:overall-accuracy} provides the average time taken to select a response for each IPD level \textit{(Time)}. Average decision-making time decreased steadily as IPD level decreased. Participants made faster decisions at lower IPD levels ($\sim$4.5s, on avg.), suggesting a reliance on gross similarities (or disimilarities) between the two images when selecting a response. At higher IPD levels, participants took longer to decide ($\sim$6s, on avg.), suggesting time taken to examine finer details. On average, participants who took 6-9s to make a decision had the highest accuracy.

\subsection{Results for Mated vs. Non-Mated Impressions}
Tabs. \ref{tab:mated-res} and \ref{tab:nonmated-res} summarize results for mated and non-mated pairs, respectively.

\begin{table}[!ht]
\centering
\caption{Mated results by IPD level of probe.}
\label{tab:mated-res}
\resizebox{\columnwidth}{!}{%
\begin{tabular}{cc|cc|cc}
\cline{3-4}
\multicolumn{2}{c|}{} & \multicolumn{2}{c|}{\textbf{\% Change in Acc.}} & \multicolumn{1}{l}{} & \multicolumn{1}{l}{} \\ \hline
\multicolumn{1}{|c|}{\textbf{IPD}} & \textbf{\begin{tabular}[c]{@{}c@{}}Acc.\\ (\%)\end{tabular}} & \multicolumn{1}{c|}{\textbf{\begin{tabular}[c]{@{}c@{}}vs. \\ Prev.\end{tabular}}} & \textbf{\begin{tabular}[c]{@{}c@{}}vs. \\ Base.\end{tabular}} & \multicolumn{1}{c|}{\textbf{\begin{tabular}[c]{@{}c@{}}Cert.\\ (\%)\end{tabular}}} & \multicolumn{1}{c|}{\textbf{\begin{tabular}[c]{@{}c@{}}Time\\ (s)\end{tabular}}} \\ \hline
\multicolumn{1}{|c|}{30} & 64.1 & \multicolumn{1}{l|}{} & \multicolumn{1}{l|}{} & \multicolumn{1}{c|}{{\color[HTML]{000000} 79.8}} & \multicolumn{1}{c|}{{\color[HTML]{000000} 5.8}} \\ \hline
\multicolumn{1}{|c|}{20} & 63.3 & \multicolumn{1}{c|}{-1.3} & -1.3 & \multicolumn{1}{c|}{{\color[HTML]{000000} 79.1}} & \multicolumn{1}{c|}{{\color[HTML]{000000} 5.8}} \\ \hline
\multicolumn{1}{|c|}{15} & 54.3 & \multicolumn{1}{c|}{-14.3} & -15.4 & \multicolumn{1}{c|}{{\color[HTML]{000000} 77.7}} & \multicolumn{1}{c|}{{\color[HTML]{000000} 5.8}} \\ \hline
\multicolumn{1}{|c|}{10} & 44.0 & \multicolumn{1}{c|}{-19.0} & -31.4 & \multicolumn{1}{c|}{{\color[HTML]{000000} 76.5}} & \multicolumn{1}{c|}{{\color[HTML]{000000} 5.5}} \\ \hline
\multicolumn{1}{|c|}{5} & 29.0 & \multicolumn{1}{c|}{-34.1} & -54.8 & \multicolumn{1}{c|}{{\color[HTML]{000000} 69.7}} & \multicolumn{1}{c|}{{\color[HTML]{000000} 4.4}} \\ \hline
\end{tabular}%
}
\end{table}

\begin{table}[!ht]
\centering
\caption{Non-mated results by IPD level of probe.}
\label{tab:nonmated-res}
\resizebox{\columnwidth}{!}{%
\begin{tabular}{cc|cc|cc}
\cline{3-4}
\multicolumn{2}{c|}{} & \multicolumn{2}{c|}{\textbf{\% Change in Acc.}} & \multicolumn{1}{l}{} & \multicolumn{1}{l}{} \\ \hline
\multicolumn{1}{|c|}{\textbf{IPD}} & \textbf{\begin{tabular}[c]{@{}c@{}}Acc.\\ (\%)\end{tabular}} & \multicolumn{1}{c|}{\textbf{\begin{tabular}[c]{@{}c@{}}vs. \\ Prev.\end{tabular}}} & \textbf{\begin{tabular}[c]{@{}c@{}}vs. \\ Base.\end{tabular}} & \multicolumn{1}{c|}{\textbf{\begin{tabular}[c]{@{}c@{}}Cert.\\ (\%)\end{tabular}}} & \multicolumn{1}{c|}{\textbf{\begin{tabular}[c]{@{}c@{}}Time\\ (s)\end{tabular}}} \\ \hline
\multicolumn{1}{|c|}{30} & 68.2 & \multicolumn{1}{l|}{} & \multicolumn{1}{l|}{} & \multicolumn{1}{c|}{{\color[HTML]{000000} 78.9}} & \multicolumn{1}{c|}{{\color[HTML]{000000} 5.8}} \\ \hline
\multicolumn{1}{|c|}{20} & 63.5 & \multicolumn{1}{c|}{-6.9\%} & -6.9\% & \multicolumn{1}{c|}{{\color[HTML]{000000} 78.4}} & \multicolumn{1}{c|}{{\color[HTML]{000000} 5.8}} \\ \hline
\multicolumn{1}{|c|}{15} & 62.5 & \multicolumn{1}{c|}{-1.7\%} & -8.5\% & \multicolumn{1}{c|}{{\color[HTML]{000000} 78.1}} & \multicolumn{1}{c|}{{\color[HTML]{000000} 5.6}} \\ \hline
\multicolumn{1}{|c|}{10} & 57.4 & \multicolumn{1}{c|}{-8.1\%} & -15.9\% & \multicolumn{1}{c|}{{\color[HTML]{000000} 77.2}} & \multicolumn{1}{c|}{{\color[HTML]{000000} 5.5}} \\ \hline
\multicolumn{1}{|c|}{5} & 42.7 & \multicolumn{1}{c|}{-25.7\%} & -37.5\% & \multicolumn{1}{c|}{{\color[HTML]{000000} 71.6}} & \multicolumn{1}{c|}{{\color[HTML]{000000} 4.6}} \\ \hline
\end{tabular}%
}
\end{table}

\subsubsection{Impact of IPD on Mated vs. Non-Mated Accuracy}

For all IPD levels, participants were more accurate in discerning that two individuals were \textit{different} than that they were the \textit{same}. At the baseline IPD of 30px, participants were 6\% more accurate for mated pairs than non-mated pairs.  At 20px, there was essentially no difference in accuracy. However, at lower IPDs, non-mated accuracy was again higher---participants were 15\% more accurate on non-mated pairs for an IPD of 15px, 31\% for 10px, and 47\% for 5px.

\subsubsection{Impact of IPD on Mated vs. Non-Mated Certainty}
In general, the percentage certainty was relatively similar across IPDs for mated and non-mated pairs. One interesting point is that at the lower IPDs (15px, 10px, 5px), participants were slightly more certain of their decisions for non-mated pairs than for mated pairs. But at higher IPDs (30px, 20px), they were slightly more certain of their decisions for mated pairs than non-mated pairs. 

The fact that certainty was so similar for mated and non-mated pairs is curious, given the significant difference in accuracy. At an IPD of 5px, for example, participants had about the same level of certainty for mated and non-mated pairs even though they were 47\% more accurate for the non-mated pairs.

\subsubsection{Impact of IPD on Mated vs. Non-Mated Time}
There was no clear trend in decision time based on pair type.

\subsection{Automated Results}
Finally, we evaluated the accuracy of an AFR system on the same impressions evaluated by the human participants. For each impression, we used ArcFace to generate similarity scores on a scale of -1 to +1. In order to determine accuracy, we used the 1-in-10k threshold (as used in \cite{frvt-demographic, gbekevi2023analyzing, krishnapriya2022analysis, krishnapriya2020issues}), which corresponds to a score of 0.356. Impressions with scores at or above the threshold were labeled as ``mated'', and those below were labeled as ``non-mated''. Tab. \ref{tab:arcface-results} provides overall accuracy, as well as accuracy for mated / non-mated pairs separately and the percent difference between the two \textit{(\% Diff. in M-NM Acc.)}. 

\begin{table}[!ht]
\centering
\caption{ArcFace results by IPD level of probe.}
\label{tab:arcface-results}
\resizebox{\columnwidth}{!}{%
\begin{tabular}{c|c|c|c|c|}
\cline{2-5}
\multicolumn{1}{l|}{} & \multirow{2}{*}{\textbf{\begin{tabular}[c]{@{}c@{}}Overall\\ Acc. (\%)\end{tabular}}} & \multirow{2}{*}{\textbf{\begin{tabular}[c]{@{}c@{}}Mated\\ Acc. (\%)\end{tabular}}} & \multirow{2}{*}{\textbf{\begin{tabular}[c]{@{}c@{}}Non-Mated\\ Acc. (\%)\end{tabular}}} & \multirow{2}{*}{\textbf{\begin{tabular}[c]{@{}c@{}}\% Diff. in \\ M-NM Acc.\end{tabular}}} \\ \cline{1-1}
\multicolumn{1}{|c|}{\textbf{IPD}} &  &  &  &  \\ \hline
\multicolumn{1}{|c|}{30} & 96.9 & 100 & 93.8 & 6.5 \\ \hline
\multicolumn{1}{|c|}{20} & 98.4 & 100 & 96.9 & 3.2 \\ \hline
\multicolumn{1}{|c|}{15} & 100 & 100 & 100 &  0 \\ \hline
\multicolumn{1}{|c|}{10} & 95.3 & 96.9 & 93.8 & 3.3 \\ \hline
\multicolumn{1}{|c|}{5} & 26.6 & 18.8 & 34.4 & 58.8 \\ \hline
\end{tabular}%
}
\end{table}

For IPDs of 30-10px, automated performance far surpassed human ability. However, for the 5px IPD, ArcFace was actually \textit{worse} than humans. For non-mated pairs at 5px IPD, ArcFace was 21.5\% less accurate than humans in correctly discerning identity mismatches. For mated pairs at 5px IPD, ArcFace was 46.3\% less accurate than humans in correctly discerning identity matches. 

\section{Conclusion \& Discussion}\label{conclusion} 

In this work, we designed and implemented a test aimed at understanding how untrained humans' ability to correctly determine identity matches and mismatches in image pairs changes as the interpupillary distance of one image decreases. In analyzing the results of this test, we determined not only how changing IPD impacted accuracy, but also how it related to decision-making certainty and time. We also found it crucial to parse the results with respect to the \textit{type} of image pair---i.e., having the same identity or different identities---as this factor had significant impact with respect to IPD. Our findings are summarized as follows.

\textbf{As IPD decreased, recognition accuracy decreased.}
Overall accuracy for image pairs containing a 30px IPD image was 66.2\% vs. only 35.9\% for pairs containing a 5px IPD image. The decrease in accuracy was consistent as IPD decreased. For pairs containing an image with $\leq$ 10px IPD, accuracy was at or below chance. This finding aligns with those in \cite{norell2015effect, keval2008can}.

\textbf{As IPD decreased, decision-making certainty decreased.}
Overall certainty for image pairs containing a 30px IPD image was 79.3\% vs. only 70.7\% for those with a 5px IPD image. The decrease in certainty was consistent as IPD decreased, though there was a much larger drop in certainty at 5px IPD versus drops between 30-10px IPD. This finding aligns with that of \cite{norell2015effect}.

\textbf{As IPD decreased, decision-making time decreased slightly.}
Average decision-making time for image pairs containing a 30px IPD image was 5.8s versus only 4.4s for those with a 5px IPD image. The most accurate decisions corresponded to a time of 6-9s. While this may seem like a relatively short time in which to make an identity decision, previous work in which participants were given either 2s or unlimited time to view a pair before deciding showed that there was not a substantial difference in accuracy between the two groups \cite{o2012comparing,phillips2012comparing}.

\textbf{As IPD decreased, humans were more accurate at discerning identity mis-matches than identity matches.}
At 30px IPD, human accuracy only differed by 6\% for mated vs. non-mated image pairs. At 5px IPD, accuracy for mated vs. non-mated pairs differed by 47\%. This result may be understood with respect to the phenomenon that, when attempting to recognize unfamiliar faces, humans rely more on external facial features (hair, ears, contour) than internal facial features (eyes, nose, mouth) \cite{latif2022importance, Bonner2003, liu2013development, want2003recognizing}. Recall that the mated / non-mated pairs were selected for difficulty \textit{based on algorithmic judgment}. Previous works show that external face features are not necessarily accessible to algorithms \cite{Rice2013, kumar2009attribute}. So, if humans presented with the 5px IPD images defaulted to relying on external features (as internal features were unavailable at that resolution), they may have been able to home in on differences in hair / ears / contour that made it easier to correctly determine an identity mis-match. (For example, it may have been more readily apparent that two sets of ears were not the same.) For the higher-IPD images, this phenomenon was less impactful because humans may have been confused by apparent similarities / dissimilarities in internal features.

\textbf{For very low IPD, humans were more accurate than a state-of-the-art recognition algorithm.}
At 5px IPD, overall human accuracy was 35\%, while ArcFace accuracy was 26.6\%. This result aligns with previous findings that when internal facial features are unclear / obscured, human reliance on external features can uncover discerning information that may not be perceivable by algorithms \cite{Rice2013, kumar2009attribute}. 

Altogether, our findings suggest that the resolution of a face image, as measured by IPD, is a \textit{crucial} factor in humans' ability to make reliable identity decisions.

This work establishes a baseline for innate human face recognition ability at low resolutions, showing that accuracy declines consistently as IPD decreases, dropping to chance levels around 10px IPD and below chance at 5px IPD. Interestingly, despite reduced accuracy, participants maintained high certainty in their answers. Notably, humans outperformed a state-of-the-art face recognition system at the lowest IPD level. Future research should evaluate expert reviewers' performance, the impact of combined image quality issues, learning / training effects over time, demographic influences, and human-algorithm collaboration to optimize identification processes in operational scenarios.

\section{Ethical Statement}
The test design and methodology used in this study were approved by the Institutional Review Board (IRB) at the Florida Institute of Technology. All participants provided informed consent, and no potential harms were identified. The images used in the study, and presented in this paper, are part of an academically available dataset, and do not contain any personally identifiable information (PII) of the subjects.

This research aims to contribute to a deeper understanding of the impact of image quality factors in operational face recognition scenarios that involve collaboration between humans and automated systems. By examining the role of these factors in human decision-making within face recognition systems, we hope to support the development of more transparent and effective operational processes.

{\small
\bibliographystyle{IEEEtran.bst}

}

\end{document}